\documentclass{llncs}

\usepackage{amsmath,graphicx}
\usepackage{amssymb}
\usepackage{amsfonts}
\usepackage{booktabs}
\usepackage{subfigure}
\usepackage{longtable}
\usepackage{caption}
\usepackage{amsmath}
\usepackage{cite}
\newcommand{\comment}[1]{}

\begin{document}

\title{A Mixture of Views Network with Applications to the  Classification of Breast Microcalcifications}


%
\titlerunning{?}  
%
\author{Yaniv Shachor\inst{1} \qquad Hayit Greenspan\inst{2} \qquad Jacob Goldberger\inst{1}}


%

\tocauthor{?}

 \institute{Faculty of Engineering, Bar-Ilan University, Ramat-Gan, Israel\\ \and
 Department of Biomedical Engineering, Tel Aviv University, Tel Aviv, Israel.\\
 }

\maketitle              

\begin{abstract}

In this paper we examine  data fusion methods for multi-view data classification. We present a  decision concept which explicitly
takes into account the input multi-view structure, where  for each case there is a different subset of relevant views.
This  data fusion concept, which we dub  Mixture of Views, is  implemented by a special purpose neural network architecture.
   It is  demonstrated on the task  of  classifying  breast microcalcifications as benign or malignant based on
    CC and MLO mammography views.
 The single view decisions are combined by a data-driven decision, according to the relevance of each view in a given case,  into a global decision. 
 The method is evaluated on a large multi-view dataset extracted from the standardized digital database for screening mammography (DDSM). The experimental results show that our  method outperforms previously suggested fusion methods.

\end{abstract}

\section{Introduction}
Over the years, research has been actively seeking the best ways to integrate
 data from multiple sources. Data fusion can be performed at the raw data, feature level and decision level. 
Feature-level fusion involves concatanating the features extracted from each view as an input to a decision system.
Decision  level  fusion  is based on  averaging  the decisions obtained from each sensor independently.

In the medical field, multi view data fusion has been researched for number of applications. Prasoon et al. \cite{prasoon} combined features from orthogonal patches in order to segment knee cartilage. Setio et al. \cite{ginneken} used decision level fusion on orthogonal patches in a pulmonary nodule detection CAD. 
The current study deals with  multi-view mammography image information fusion. In this task, information from images acquired from different angles is aggregated for automatic classification of breast microcalcifications (MC) as benign or malignant.
  Recent studies (e.g. \cite{Samulski2011}\cite{Velikova2012}) have confirmed the superior
performance of a multi-view CADx system over its single-view counterpart. 
Bekker et al.  \cite{Bekker_TMI_2015} proposed averaging the two  view-level decisions.
There is, however, a structure specific to this problem
that is not explicitly modeled by  equal weight averaging or even by  a weighted average with fixed weights.
 Screening mammography typically involves taking two views of the breast, from above and from an oblique,
 since  a 3D pathology indication is not always clearly observed in a single 2D image.
      Thus an expert radiologist's  diagnosis of malignance is based on evidence that is not necessarily  clearly seen in both views.
    \\In this study we introduce a neural network (NN) architecture that explicitly takes into account the multiple-view structure of the problem by
integrating view level decision making.
     We present an automatic data-driven strategy that finds which view  conveys more relevant information for  clinical decision making.
Instead of  simple averaging of the view-level decisions, we train a `gating' network that decides  the best way
to average the view-level decisions for each case.
Our method is related to the well-known Mixture of Experts (MoE) model \cite{JacobsJordanNowlanEtAl91}. 
Unlike classical MoE which divides the task among a set of experts in an unsupervised way, in MoV each expert is associated with a sensor (or a view). We describe a neural network architecture that implements the MoV concept.

\begin{figure}[ht]
\center
\includegraphics[width=8cm]{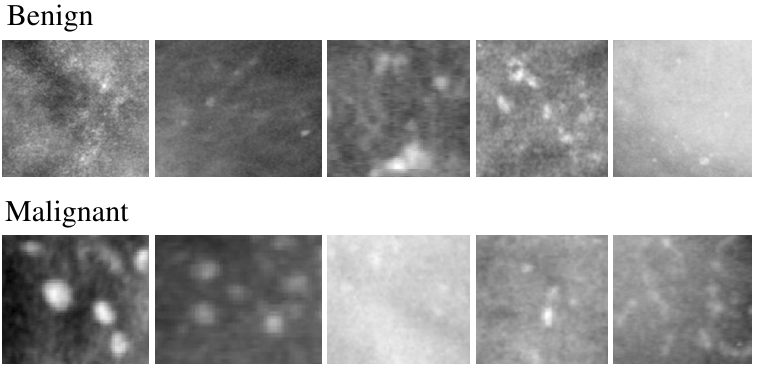}
\caption{Examples of benign and malignant MC clusters (from
the DDSM dataset). ROI sizes range from $2mm \times  2mm$ to
$6mm \times 6mm$.	}
\label{fig:samples}
\end{figure}

To evaluate the  method we use  the labeled multi-view mammogram  dataset DDSM \cite{DDSM}.
It contains MCs location in both views marked by experts and we also have the biopsy results,
 showing whether the abnormities were benign or malignant.
    Experiments were performed on pairs of CC+MLO views extracted from the DDSM dataset. Fig. \ref{fig:samples}
    shows benign and  malignant examples from the DDSM dataset.
The results  when applying the MOV to the DDSM dataset show  that using our  approach outperforms   previously suggested  fusion methods.

\section{ The Mixture of Views Network architecture}

In this study we deal with a classification task where the  features are obtained from multiple sensors.
Our goal is to construct a neural network architecture which is aware of this input structure and  takes advantage of  it to improve  classification performance.  In this section we first describe the probabilistic
framework we use to model the multi-view data fusion classification  and then
derive a training algorithm that simultaneously finds the parameters of each view-based classifier and the
 parameters of a  gating network that decides which view is more relevant for a given input feature set.
 Hereafter we use the terms sensor and view interchangeably.

Assume a feature vector $x$ is a concatenation  of $m$  components $x=(x^1,...,x^m)$ such that each component is obtained from a different sensor.  Each sensor can provide a different number of features. The standard way of classifying an object based on features from  multiple views is to concatenate the view-level inputs and using the concatenated vector as an input to a standard neural-network  classifier.

  A model based on concatenation, however, does not take into account the structure of the  system, as a fusion of several views. Each view has enough information to make its own reasonable prediction and, in principle, each component can be used alone for the classification task.
  In addition not all the views are equally relevant in each case.
   We propose a model that computes each view-level prediction separately and then combines them to form the final estimation.  In other words, in our approach we concatenate the  view-level decisions instead of the view-level features.
  We also analyze the features provided by different views and combine the predictions dynamically, according to the relevance of the view in each case.

Assume we are given a $k$-class  classification problem with labels denoted by $1,...,k$ and  input $x$
 composed of $m$ view-level components denoted by $x^1,...,x^m$.
For the classification task we use a neural network that combines  the view-level decisions.
 Let $\theta = \{\theta_g,\theta_{1},\theta_{2},\dotsm,\theta_{m}\}$ be the  model parameters, where $\theta_i$ are the parameters
 of the network that performs a classification  based solely on the component $x^i$  and $\theta_g$ is the parameter-set of a gating network that
 computes a data-driven distribution over all the views. The final decision is derived by computing a  weighted average of the view-level decisions where the weights are provided by the gating networks.

The network  has three main components:
a set of neural-networks where each one renders a decision based solely on a single sensor, a gate neural-network that defines those sensors whose individual sensor-based opinions
are trustworthy, and  a weighted sum of experts,
where the weights are the gating network outputs.

 In the MoV model the probability of input $x$ being labeled as $c\in \{1,...,k\}$ is:
\begin{equation}
  \label{MoV}
p(y =c|x;\theta) = \sum_{i=1}^m p(i|x;\theta_g) p(y=c|x^i;\theta_i).
\end{equation}
We can view the model as a two-step process that
produces a decision $y$ given an input feature set $x$. We first use
the gating function to select the view that conveys the relevant information for decision making and then apply the corresponding  network
to determine the output label. 

As stated above, the MoV model can be viewed as an instance of Mixture of Experts (MoE) modeling.
  The MoE approach  was  introduced more than twenty years ago \cite{JacobsJordanNowlanEtAl91},
 and combines the decisions of several experts, each of
which specializes in a different part of the input space.
 The MoE model is based on the ``divide and conquer" paradigm, which solves complex problems by dividing them into simpler ones and combine their solutions to solve the original task.
 The  model allows the individual experts to specialize on
smaller parts of a larger problem, and it uses soft partitions
of the data implemented by the gate.  In the general setup of MoE all experts are exposed to all the features  and the goal is to cluster the feature space and associate each cluster with an expert classifier in an unsupervised manner.
  By contrast, in the MoV model there is a predefined partitioning of the set of features according to the sensors used to measure them
   and each expert is specialized in  making decisions based solely on the features of the corresponding sensor.

\comment{As illustrated in Fig.\ref{fig:moe3} The ME model has 3 main components: several experts networks,gate network and probabilistic model to combine the experts and the gate. Each expert receives the input features x and produces prediction.The gate network receive x and produces weights in which the final prediction use as a weighted sum of all experts predictions.
The number of expert models is a hyper-parameter of the model. This setup allow each expert model to specialize on subset of the input data,creating nonlinear separation in the input space. The gate model is responsible for weighting the experts' predictions on the current input,according to their area of expertise.}

We next describe the training procedure. Assume we are given $n$ feature vectors  $x_1,...,x_n$ with corresponding labels $y_1,...,y_n \in\{1,...,k\}$.
Each input vector $x_i$ is composed of $m$ components  $x_i^1,...,x_i^m$ collected from the $m$ sensors.
The log-likelihood function of the model parameters is:
\begin{equation}
L(\theta)  = \sum_t \log p(y_{t}|x_t;\theta)  = \sum_t \log (\sum_i p(i|x_t;\theta_g) p(y_t|x_t^i;\theta_i)). 
\label{likelihood}
\end{equation}

    To find the network parameters we can maximize the likelihood function using  the standard  back-propagation algorithm.
     It can be easily verified that the back-propagation equation for the parameter set  of the $i$-th expert is:
\begin{equation}
\frac{\partial L}{\partial {\theta}_{i}} = \sum_{t=1}^n w_{ti} \cdot \frac{\partial} {\partial \theta_{i}}
 \log   p(y_t|x_t^i;\theta_i)
\label{dmoefder1}
\end{equation}
such that $w_{ti}$ is the posterior distribution of the gating decision:
\begin{equation}
{w}_{ti}=  p( i|x_t,y_t;\theta) = \frac { p(y_t|x_t^i;\theta_i)p(i|x_t;\theta_g)}{p(y_t|x_t;\theta)}.
\label{sestep}
\end{equation}
In a similar way,  the  back-propagation equation for the parameter set of the gating network   is:
\begin{equation}
\frac{\partial L}{\partial {\theta}_{g}} = \sum_{t=1}^n  \sum_{i=1}^m w_{ti} \cdot \frac{\partial }{\partial \theta_g} \log
 p(i|x_t;\theta_g).
\label{dmoefder2}
\end{equation}
\comment{
\begin{figure}[ht]
\includegraphics[width=8cm]{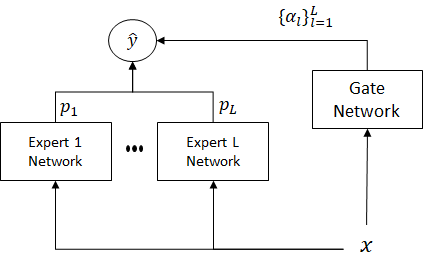}
\caption{ Mixture of Views (MoV) architecture.}
\label{fig:mov}
\end{figure}
}

The likelihood score (\ref{likelihood}) is focused on the performance of the compound network. It does not, however, explicitly encourage each view-level network to obtain the optimal decision based on features of the corresponding view. One strategy to overcome this issue is  to train initially each view-level network separately and then to use  the learned parameters as initial values for the compound network.  We have found that this approach, which  is based on injecting information to the system via parameter initialization, does not work well since the network tends to forget its initialization after a few training iterations.
Rather, we use a modified likelihood score:
\begin{equation}
\label{eq:loss}
 L(\theta) + \lambda \sum_i L_i(\theta_i)
\end{equation}
such that $L(\theta)$ is the usual likelihood score (\ref{likelihood}) and
$
L_i(\theta_i) =  \sum_t \log p(y_{t}|x_t^i;\theta_i)
$
is the likelihood score where we only use the network corresponding to the $i$-th view for classification.
 The parameter $\lambda$  controls the relative importance of the view-level and integrated decisions and can be tuned using cross-validation.

\comment{
We can view the gating decision  as a hidden random variable. We can thus use the EM algorithm \cite{twenty} to optimize the likelihood function
(\ref{likelihood}) instead of the back-propagation procedure described above. In the E-step we estimate the relevant sensor in a way similar to equation
(\ref{sestep}) and in the M-step we  train each of the view-level networks and the gating network separately.
However, there are several drawbacks to training neural  networks with the EM-based approach. The EM algorithm is
a greedy optimization procedure that is notoriously known to get stuck at local optima. Another
issue related to  combining neural networks and EM direction is computational cost. The framework
requires training a neural network on each iteration of the EM algorithm. This is a costly computation for large networks, and does not scale well. In addition,  since both EM and the neural network optimization use iterative algorithms, the training schedule is unclear.
Therefore, we prefer to use a neural-network based optimization procedure rather than the EM algorithm.
}
\comment{
\begin{figure}[thpb]
\begin{center}
{\centering \includegraphics[scale=0.7]{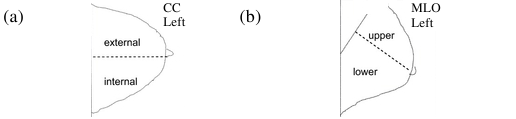} }
\end{center}
\caption{Illustration of the two image views, (a) CC view and (b) MLO view.}
\label{CCnMLO}
\end{figure}
}
\section{Multi-view Classification of Breast Microcalcifications }
In this section we demonstrate the MoV method in the task of classifying  breast microcalcifications as benign or malignant, based on two mammography views.
A screening mammographic examination usually consists of four images, corresponding
to each breast scanned in two views: the mediolateral oblique (MLO) view and the craniocaudal (CC) view. The MLO projection is taken in a $45^{\circ}$ angle and shows part
of the pectoral muscle. The CC projection is a top-down view of the breast.    Both views are included in the diagnostic procedure.
 When reading mammograms, radiologists judge whether or not a malignant lesion is present by examining both views and breasts.
 In an expert diagnosis procedure, the expert looks at each of the views separately, and delivers one final assessment.

\begin{figure}[ht]
\center
\includegraphics[width=7cm]{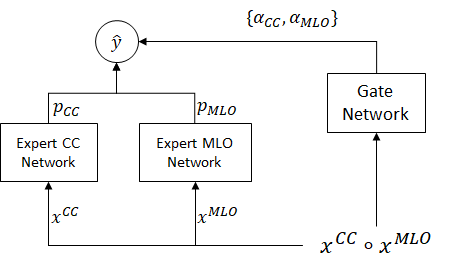}
\caption{ MoV network architecture for multi-view classification of breast microcalcifications.}
\label{fig:movccmlo}
\end{figure}

We next apply the MoV method presented above for classification of breast microcalcifications as benign or malignant. In this task, let $\{x^{cc}, x^{mlo}\}$ denote  the extracted features from the CC and MLO views, respectively. The MoV network for this task is:
\begin{equation}
p(c|x;\theta) = \sum_{i\in{\{cc,mlo\}}}p(i|x^{cc} \circ x^{mlo};\theta_g) p(c|x^{i};\theta_{i})
\end{equation}
such that $c$ is either benign or malignant.
The network is illustrated in
  Fig. \ref{fig:movccmlo}. In the next section we show empirically that the MoV network yields better classification results
than networks that are based on a single view and better results than other strategies that  combine information from  the two views.

\section{Experimental Setup}

{\bf Dataset and features. }
The empirical evaluation was based on  the DDSM dataset \cite{DDSM} which provides a high number of annotated mammograms with a biopsy-proven diagnosis.
We extracted ROIs that contained clusters of MCs for which a proven pathology had been found.
In real-life diagnosis procedures, radiologists pay attention to the density of the breast, which can be a good predictor of a woman's breast cancer risk. We assumed that categorizing the mammograms based on their density was necessary  to compare the different features and classifiers in a more objective
way.
 We separated the mammograms into two different tissue-density categories and studied them individually:  fatty tissues (ratings 1 and 2),  and dense tissues (ratings 3 and 4).
        We considered classifying the  fatty and dense breast densities as two separate tasks. The density type for each case was a parameter supplied by an expert as part of the DDSM.
We chose patients in the DDSM dataset that had both CC and MLO views  to test our model.
 Our dataset was comprised of 1410 clusters (705 of CC, and 705 of MLO), of which 372 were benign and 333 were malignant.

Feature vectors $x_{cc}$ and $x_{mlo}$ were extracted from the CC and MLO views, respectively.
 Following \cite{Bekker_TMI_2015}, texture  features were extracted from the Curvelet coefficients at intermediate scales (in our study, two scales), and included the four features mentioned in \cite{rotation_invariant} for each scale, with three additional features: entropy, skewness and kurtosis. Overall, each extracted ROI was represented by 14 features.
\comment{Many other texture features that can be used for mammography analysis have been reported in the literature, e.g.
 GLCM \cite{haralick},  (GLRLM) \cite{GLRLM} \cite{Texture_run_length}, Gabor filters \cite{DaugmanGabor} and features that are based on the wavelet transform. Using the Curvelet features we obtained the best results. Since this is not the focus of this work, we do not describe here classification results based on the alternative features.}

{\bf Training procedure.} We used a 10-fold cross validation setup. In this setup, there is
 complete isolation of the test set from the train set. Each
fold was only used for testing and never for training. We thus
ensured that no bias was introduced.
In addition, 10\% from each training set was used as a validation set in order to optimize the model hyper-parameters, according to the mean results on the validation sets.
Using the features described in  the previous section, the size of the input feature set is 28 (14 features for each view).
The features of each view  were inserted into the expert NN. In addition, all the features were inserted into the gate NN, to receive the weights for the experts` predictions.
The expert NN has 2 hidden layers comprised of 24 neurons each.
The gate NN has 2 hidden layers, comprised of 3 neurons each.
We used ReLU non-linear activations between the layers and dropout with parameter 0.5.
We used Eq.  (\ref{eq:loss}) as a target function for optimization where $\lambda = 1$ gave the best results.
The objective function  was maximized using the Adam optimizer \cite{Adam}.
The network trained over 500 epochs with reduction of the learning rate on the  loss plateau and early stopping.


\begin{figure}[h]
\subfigure{\includegraphics[width=6cm]{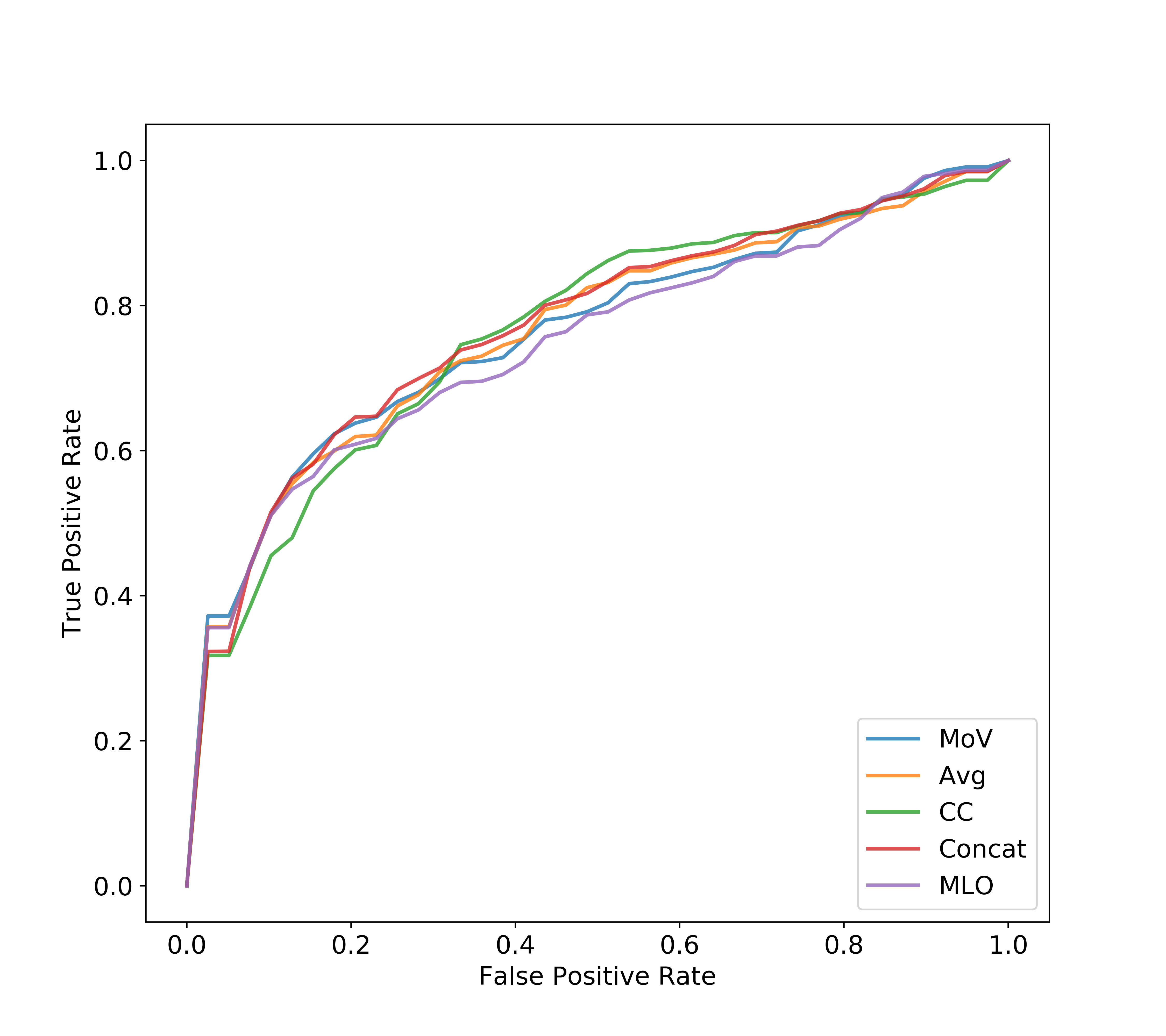}}
\subfigure{\includegraphics[width=6cm]{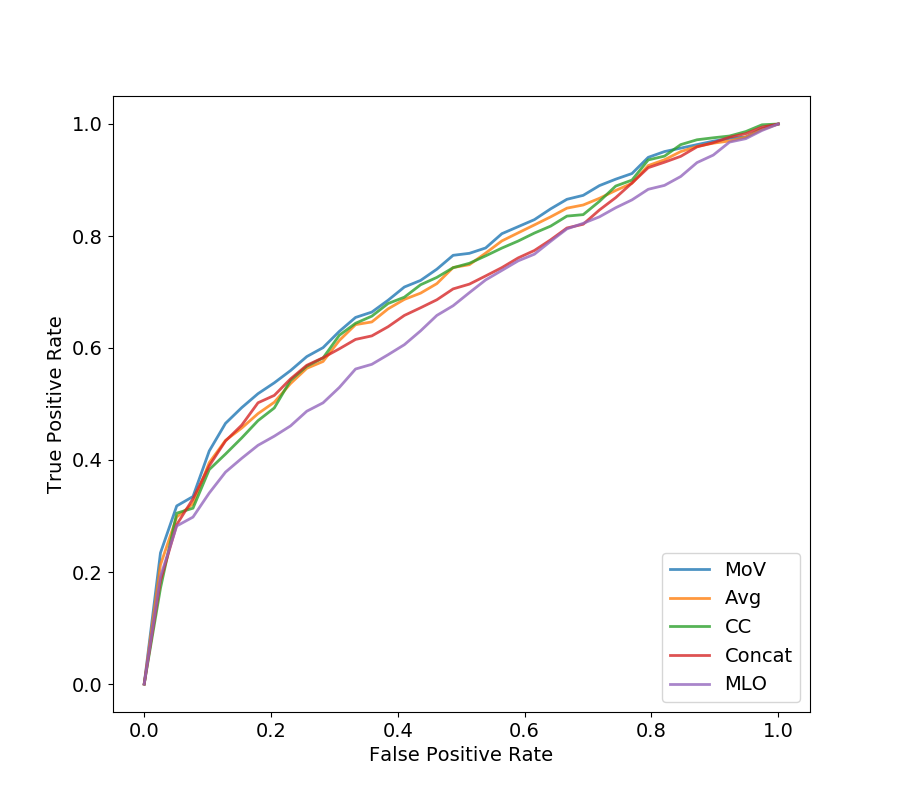}}
\caption{ROC curves for the fatty (left) and dense (right) tissues.}
\label{fig:roc}
\end{figure}

\begin{table}[th]
\centering
\caption{Classification  results (benign vs. malignant) for fatty (left) and dense (right) breast tissues.}
\label{fatty}
\begin{tabular}{|l|c|c|c|}
\hline
Method       & Accuracy       & F-measure      &AUC\\
\hline
CC      & 0.704          & 0.630  &0.695         \\
\hline
MLO    & 0.698          & 0.583         &0.682 \\
\hline
Avg \cite{Bekker_TMI_2015} \cite{ginneken}          & 0.709          & 0.620          &0.699 \\
\hline
concat \cite{prasoon}  & \textbf{0.718} & 0.624          &0.704\\
\hline
MoV          & \textbf{0.718} & \textbf{0.632} &\textbf{0.708} \\
\hline
\end{tabular}
\hspace{0.5cm}
\label{dense}
\begin{tabular}{|l|c|c|c|}
\hline
Method       & Accuracy       & F-measure     &AUC  \\
\hline
CC      & 0.643          & 0.580          &0.639 \\
\hline
MLO     & 0.629          & 0.520         &0.622\\
\hline
Avg \cite{Bekker_TMI_2015} \cite{ginneken}          & 0.646          & 0.573 &0.641          \\
\hline
Concat \cite{prasoon} & 0.642 & 0.568  &0.637        \\
\hline
MoV          & \textbf{0.665} & \textbf{0.596} &\textbf{0.661} \\
\hline
\end{tabular}
\end{table}

{\bf Experimental evaluation.}
We compared the MoV method to several models: 
a network based on the single-view (CC or MLO) features as input,
  and a network that averages the decisions based on  CC and  MLO \cite{Bekker_TMI_2015} \cite{ginneken}. This model can be viewed as a degenerated version of MoV where the data-driven gating network is replaced by a simple averaging.  We denote this model as Avg.
We also implemented a network that  concatenates  CC and MLO features as input, as in \cite{prasoon}, denoted as  Concat.

We optimized each model's hyper-parameters using a validation set, to conduct a fair comparison.
We used Receiver Operator
Characteristic (ROC) curves with accuracy, Area Under Curve (AUC) and F-measure metrics to evaluate the algorithms' performance.
Due to the differences in nature between the two types of breast tissue (Fatty/Dense), a different network architecture was trained for each type, and therefore each breast tissue category was evaluated separately.
 Table 1 shows the  classification results for
the two types of breast tissue (Fatty/Dense), and Fig. \ref{fig:roc} shows the corresponding ROC curves. The results are the 10-fold test set classification average, computed over several experiments.
As can be seen from the tables, using the  MoV method yielded the best results over the dense samples in all metrics and over the fatty samples, had the best results on the F-measure and AUC and had the best accuracy on a par with the Concat  network.

We performed DeLong test \cite{delong} comparing the MoV model paired with each of the baseline models. The input to the DeLong test consisted of predictions from the 10-fold cross validation with the corresponding labels. The DeLong test examined the null hypothesis that both methods have the same AUC. On the dense data, all the hypotheses
were successfully rejected with p-value $< 0.05$. On the fatty data we have not seen significance in the results. We assume it is because the fatty data has half of the size of the dense dataset.

To  conclude, 
in this study we addressed the problem of fusing several data sources for a classification task.
We proposed a network architecture  that is explicitly aware that the  input data are provided by multiple sensors. We demonstrated the
performance of the MoV method on the classification of
breast microcalcifications into benign and malignant
given multi-view mammograms. We showed that the MoV
architecture yields improved performance. In the algorithm presentation we focused
on the a two-view mammography, namely CC
and MLO. Our method, however, can be easily extended to
mammography with more than two views. In addition, the ROI features could be extracted automatically using transfer learning methods. In the future
we plan to investigate the applicability of the  method to
other medical imaging tasks with multiple views and/or multiple scans.

%
%
%
\bibliography{main}
\bibliographystyle{splncs03}

\end{document}